\pdfoutput=1

\documentclass[11pt]{article}

\usepackage{acl}

\usepackage{times}
\usepackage{latexsym}

\usepackage[T1]{fontenc}

\usepackage[utf8]{inputenc}

\usepackage{microtype}

\usepackage{url}
\usepackage{tipa}

\usepackage{amsmath}

\usepackage{linguex}

\usepackage{tipa}

\usepackage{todonotes}
\makeatletter
\newcommand*\iftodonotes{\if@todonotes@disabled\expandafter\@secondoftwo\else\expandafter\@firstoftwo\fi}  
\makeatother

\newcommand{\citeposs}[1]{\citeauthor{#1}'s (\citeyear{#1})}

\usepackage{cleveref}
\crefname{section}{\S}{\S\S}
\Crefname{section}{\S}{\S\S}
\crefname{table}{Table}{}
\crefname{figure}{Fig.}{Figs.}
\crefname{algorithm}{Algorithm}{}
\crefname{algorithm}{Algorithm}{}
\crefname{line}{Line}{}
\crefname{appendix}{App.}{}
\crefformat{section}{\S#2#1#3}
\crefname{thm}{Theorem}{}
\crefname{cor}{Corollary}{}
\crefname{prop}{Proposition}{}
\crefname{def}{Definition}{}

\usepackage{graphicx}
\graphicspath{ {./images/} }

\usepackage{multirow}
\usepackage{lipsum}
\usepackage{xcolor}
\usepackage{color,soul}

\title{Does BERT \textit{really} agree ?\\Fine-grained Analysis of Lexical Dependence on a Syntactic Task}

\newcommand{\lattice}{\normalfont \text{\textipa{@}}}
\newcommand{\unipi}{\normalfont \text{\textipa{B}}}

\author{Karim Lasri$^{\lattice,\unipi}$ Alessandro Lenci$^{\unipi}$ Thierry Poibeau$^{\lattice}$ \\
$^{\lattice}$Lattice (\'Ecole Normale Supérieure-PSL, CNRS, U. Sorbonne Nouvelle)~\; \\ 
~$^{\unipi}$University of Pisa~\;  \\
  \texttt{\href{mailto:karim.lasri@ens.psl.eu}{karim.lasri@ens.psl.eu}}~\;~  \\ \texttt{\href{mailto:alessandro.lenci@unipi.it}{alessandro.lenci@unipi.it}}~\;~ \texttt{\href{mailto:thierry.poibeau@ens.psl.eu}{thierry.poibeau@ens.psl.eu}}
}

\begin{document}

\maketitle

\begin{abstract}

Although transformer-based Neural Language Models demonstrate impressive performance on a variety of tasks, their generalization abilities are not well understood. They have been shown to perform strongly on subject-verb number agreement in a wide array of settings, suggesting that they learned to track syntactic dependencies during their training even without explicit supervision. In this paper, we examine the extent to which BERT is able to perform lexically-independent subject-verb number agreement (NA) on targeted syntactic templates. To do so, we disrupt the lexical patterns found in naturally occurring stimuli for each targeted structure in a novel fine-grained analysis of BERT's behavior. Our results on nonce sentences suggest that the model generalizes well for simple templates, but fails to perform lexically-independent syntactic generalization when as little as one attractor is present. 

\end{abstract}

\section{Introduction}


Every English speaker would judge as grammatical the sentences in \ref{ex1a}-\ref{ex1b}, but not those in  \ref{ex1c}-\ref{ex1d}, despite that they are all meaningless:

\begin{small}
\ex.\label{ex1}
\a.\label{ex1a} Colourless green \textcolor{blue}{ideas} \textcolor{red}{sleep} furiously.
\b. \label{ex1b} Colourless green \textcolor{blue}{ideas} that cook the door \textcolor{red}{sleep} furiously.
\b. \label{ex1c}*Colourless green \textcolor{blue}{ideas} \textcolor{red}{sleeps} furiously.
\b. \label{ex1d}*Colourless green \textcolor{blue}{ideas} that cook the door \textcolor{red}{sleeps} furiously.

\end{small}

\noindent{}At least since \citet{Chomsky:1957}, data like this has been taken as evidence that natural language grammars contain abstract syntactic rules that (i) are independent of the meaning of lexical items and (ii) obey hierarchical, rather than linear constraints. Number agreement (henceforth NA) between the subject (the \textbf{cue}) and the verb (the \textbf{target}) of the same clause in English is one of such rules \cite{Corbett2003AgreementTA}. In fact, \ref{ex1d} is ungrammatical, even though the closest noun \textit{door} (typically referred to as \textbf{attractor}) has the same number as \textit{sleeps}, because the noun belongs to an embedded relative clause. These NA properties have made it one of the preferred test beds to investigate the ability of neural language models (NLMs) to learn abstract, hierarchical syntactic structures \cite{linzen-etal-2016-assessing, gulordava-etal-2018-colorless, marvin-linzen-2018-targeted, goldberg, bacon, lakretz-etal-2019-emergence}. Although recurrent and transformer-based NLMs have been shown to possess syntactic abilities on the task, their nature is not fully understood \cite{baroni2}.


Can NLMs really perform lexically-independent number agreement, regardless of the syntactic structure? 
To answer this question, we test BERT \cite{devlin-etal-2019-bert} against the NA task while controlling both the syntactic constructions and the meaningfulness of the stimuli presented to the model. 

 
Our experiments provide two main findings. Contrarily to previous observations that BERT performs fairly well on \citeposs{gulordava-etal-2018-colorless} syntactically well-formed but meaningless sentences \cite{goldberg}, we show that its generalization abilities are not lexically-independent on syntactic constructions where an attractor is present\footnote{As in \ref{ex1b} and \ref{ex1d} above}. 
Though the model has been previously shown to ignore attractors belonging to an embedded clause independent of that containing the target \cite{goldberg}, we further provide insights on this lexical dependence that reveal the limitations of the model's abilities. Our experiments rather show that the model is actually sensitive to the presence of attractors when semantic and lexical patterns are disrupted in its input sentence.

\section{Related work}
\citet{linzen-etal-2016-assessing} first tested the ability of LSTM language models to solve the NA task, and showed that they capture syntax-sensitive dependencies given targeted supervision. A subsequent study by \citet{gulordava-etal-2018-colorless}, showed that LSTMs are able to succeed even on nonce sentences obtained by replacing the lexical content in the used stimuli while keeping the syntactic structure unchanged. This suggested NLMs can acquire grammatical competence that goes beyond meaningful lexical patterns they have seen during training on a language modeling objective. 
\citet{marvin-linzen-2018-targeted} further tested an LSTM's ability to capture syntactic dependencies on constructed pairs of meaningful manually crafted sentences, so as to test targeted syntactic constructions. Contrarily to previous studies, they showed that there was considerable room for improvement for LSTMs on some challenging syntactic structures.

\citet{goldberg} further tested BERT, a transformer-based model, against stimuli from \citet{linzen-etal-2016-assessing}, \citet{gulordava-etal-2018-colorless} and \citet{marvin-linzen-2018-targeted}. He found that BERT substantially outperforms the previously tested LSTM language models.

\citet{newman-etal-2021-refining} have recently tested generalizations beyond \citeposs{marvin-linzen-2018-targeted} data by extending the vocabulary at the target verb position. They show that though NLMs' top predictions are generally correct verbforms, the models still struggle on the NA task for infrequent verbs.
In addition to testing the effect of meaningfulness by performing replacements at all positions of the sentence similarly to \citet{gulordava-etal-2018-colorless}, we control for the syntactic constructions from \citet{marvin-linzen-2018-targeted}:
given a syntactic template, can BERT generalize to \textit{any} syntactically well-formed, but meaningless sentence? If not, when does lexical content matter? 

\section{General Setup}
\subsection{The Number Agreement Task}
The NA task consists in testing whether a model shows a preference for predictions that do not violate number agreement between a selected verb and its subject. For example, when presenting BERT with sentences \ref{ex1b} and \ref{ex1d}, we mask the token at the target position, and compare the output probabilities for \textcolor{red}{sleep} and \textcolor{red}{sleeps}. The model succeeds when it assigns a higher prediction score to the right target form.

\subsection{Datasets}
We test BERT's ability to solve the NA task using three different, but complementary datasets all consisting of sentences controlled by the syntactic templates described in \cref{agreement-structures}:

\noindent{}a) \textbf{M\&L}. This is the original dataset released by \citet{marvin-linzen-2018-targeted}, containing the syntactic constructions we use in this study. We use it to replicate \citeposs{goldberg} results as a comparison point. These sentences were designed to respect semantic constraints using a limited, but semantically controlled vocabulary.

\noindent{}b) \textbf{WIKI}. For each template in \textbf{M\&L}, 
we collected naturally occurring sentences from the Wikidumps used to train BERT, to test whether the model performs better on sequences of words it could have memorized during training. We extracted raw text from the Wikidumps using WikiExtractor\footnote{\url{https://github.com/attardi/wikiextractor}}, and collected sequences of word that corresponded to the sequence of POS tag for each template in \textbf{M\&L}. The data collection procedure is described in \ref{sec:natural-data}.

\noindent{}c) \textbf{NONCE}. For each template in \textbf{M\&L}, we generated ``nonce'', meaningless sentences keeping the syntactic structure unaffected\footnote{We release this data on \url{https://github.com/karimlasri/does-bert-really-agree}}. To do so, we replace each word in the sentence with a word of the same lexical category (and same number if applicable) using a large set of words for each POS-tag (see \cref{sec:vocabulary}), similarly to \citeposs{gulordava-etal-2018-colorless} stimuli. When a noun intervenes between the cue and the target (e.g., in condition C from \cref{agreement-structures}), it is systematically assigned a different number from the cue, in order to test attraction effects\footnote{That is whether the model succeeds despite the presence of a distractor noun between the cue and target of the agreement.}. 
These nonce sentences are meaningless, therefore they violate selectional restrictions contrarily to \textbf{M\&L}. They also differ from \citeposs{gulordava-etal-2018-colorless} stimuli as we additionally test the effect of the syntactic construction, having separate conditions for each template.  
This dataset allows us to test the extent to which the model's ability to perform the agreement on nonce sentences is dependent on their syntactic structure. Each set contains 10000 sentences, with balanced proportions of singulars and plurals, making chance level at 50\%. 


\begin{table*}
\begin{small}
\begin{center}
\begin{tabular}{ cll } 
 \hline
 \textbf{Struct. ID} & \textbf{Structure description} & \textbf{Example} \\
 \hline
 \textbf{A} & Simple agreement & The \textbf{\textcolor{blue}{boy}} \textbf{\textcolor{red}{laughs}}/*laugh \\
 \textbf{B} & In a sentential complement & The boy knows the \textbf{\textcolor{blue}{girls}} \textbf{\textcolor{red}{play}}/*plays \\
 \textbf{C} & Across a prepositional phrase & The \textbf{\textcolor{blue}{plate}} near the \underline{glasses} \textbf{\textcolor{red}{breaks}}/*break \\
 \textbf{D} & Across a subject relative clause & The \textbf{\textcolor{blue}{cat}} that chases the \underline{mice} \textbf{\textcolor{red}{runs}}/*run \\
 \textbf{E} & In a short verb phrase coordination & The \textbf{\textcolor{blue}{boy}} smiles and \textbf{\textcolor{red}{laughs}}/*laugh \\
 \textbf{F} & Across an object relative clause & The \textbf{\textcolor{blue}{mouse}} that the \underline{cats} chase \textbf{\textcolor{red}{runs}}/*run \\
 \textbf{G} & Within an object relative clause & The mouse that the \textbf{\textcolor{blue}{cats}} \textbf{\textcolor{red}{chase}}/*chases runs \\
 \textbf{H} & Across an object relative clause (\textit{no that}) & The \textbf{\textcolor{blue}{mouse}} the \underline{cats} chase \textbf{\textcolor{red}{runs}}/*run \\
 \textbf{I} & Within an object relative clause (\textit{no that}) & The mouse the \textbf{\textcolor{blue}{cats}} \textbf{\textcolor{red}{chase}}/*chases runs \\
 \hline
\end{tabular}
\caption{\label{agreement-structures}Agreement structures used in this study. These structures are taken from \citet{marvin-linzen-2018-targeted}. The cue is in blue and the target is red. For each target, we display the pair of both the correct and incorrect verb form. In structures C, D, E and H, the attractor is underlined. }
\end{center}
\end{small}
\end{table*}


\section{Experiments and Results}
\subsection{EXP. 1 -- Sensitivity to Meaning on a Syntactic Task}

In this experiment, we test whether the model's success over the NA task on \citeposs{marvin-linzen-2018-targeted} syntactic templates requires satisfying mutual semantic constraints. To do so, we compare the NA task accuracy on \textbf{M\&L} and \textbf{NONCE}. We also use \textbf{WIKI} as a comparison point, to observe whether the model succeeds better on sentences it could have memorized during training than on \textbf{M\&L}'s meaningful but unseen sentences.

\begin{figure}
    \begin{center} \includegraphics[width=\columnwidth]{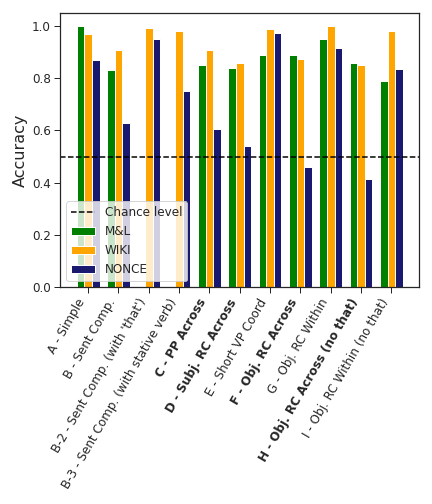}
    \caption{Accuracies on the number agreement task for the retained structures obtained by BERT Base. Templates where an attractor is present are displayed in bold. Note that conditions B-2 and B-3 were not present in the original \textbf{M\&L} stimuli}
    \label{fig:raw-accuracies}
    \end{center}
\end{figure}

The results from \cref{fig:raw-accuracies} show that even though BERT is quite robust against all templates on stimuli from \citet{marvin-linzen-2018-targeted}, it fails on some templates in \textbf{NONCE}.
Little performance reduction occurs when there is no intervening attractor (A, E, G, I), that is when the cue and target are within the same clause. This shows that the model can solve the NA task in the absence of attractors, even when there is a violation of semantic selectional restrictions. The only exception is when the cue occurs in a sentential complement (B). In the absence of the complementizer \textit{that}, the model might be perturbed by ambiguity, expecting a direct object noun (e.g., \textit{The boy knows the mathematics lessons}). Therefore, we tested two supplementary conditions: one with the overt complementizer (B-2), and another where the verb that introduces the complementizer is constrained to be a stative verb (B-3). The results confirm our hypothesis: BERT carries out the task successfully on \textbf{NONCE} when the complementizer makes the sentence syntactically unambiguous, which also suggests that the model relies on heuristics that are partly lexicalized.
On the other templates, performance drops close to chance level on \textbf{NONCE}. This means that BERT is not able to perform lexically-independent 
generalizations when the target and the cue are separated by a hierarchically embedded phrase containing an attractor noun. Interestingly, the model often performs better on \textbf{WIKI} than on \textbf{M\&L}, which suggests that memorized lexical 
patterns can help solve the task in addition to being meaningful.


\begin{figure*}[h]
    \begin{center}
    \includegraphics[width=\textwidth]{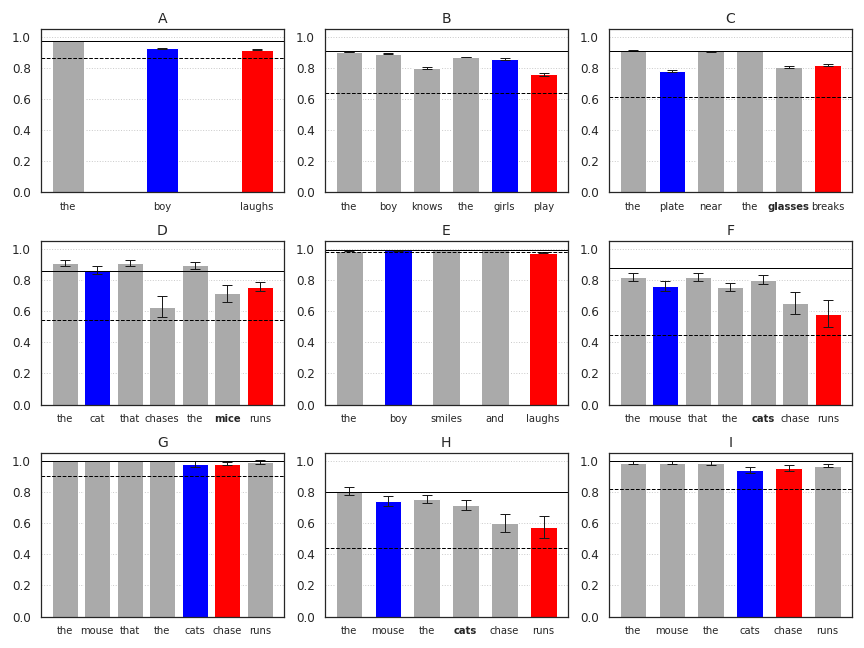}
    \caption{\label{one-borrowed-accs-fig}Accuracies on the NA task after one-word replacement. Each column represents the model's performance after intervening at the position exemplified by the word displayed in the x-axis. Attractors are represented in bold. Replacements are performed over sentences from \textbf{WIKI}. For each syntactic template, the performance on \textbf{WIKI} (continuous line) and \textbf{NONCE} (dashed line) is represented as a comparison point. The cue's replacement is represented in blue and the target's in red.}
    \end{center}
\end{figure*} 

\subsection{EXP. 2 -- Influence of One-Word Replacements}

In this experiment, we measure how performance is affected when replacing words at one position at a time in the templates, on \textbf{WIKI}. Our goal is to understand whether the performance drop observed in EXP.~1 is due to the lexical content filling specific syntactic positions in our templates. In particular, we wish to  understand whether most of the effect is due to replacing the cue, the target, the attractor (if present) or words in none of those three categories. 

The results in \cref{one-borrowed-accs-fig} show that in sentences with no attractor (A, E, G, I), one-word replacement results in low performance drops, consistently with observations from EXP. 1. When the stimuli contain an embedded phrase containing an attractor, replacing the target itself, but also words close to the target verb (in D, F and H) can significantly harm performance. The cue is linearly distant from the target in sentences with attractors, and its replacement has little impact on performance. We observe that replacing the attractor replacement also has a limited impact on the task, as templates D and H show. 
We note a general tendency that replacing closest words results in higher performance drop than replacing farther ones, including verbs in embedded clauses.
This suggests that the model's ability to deal with attractors is not due solely to hierarchical, lexically independent generalizations acquired during training. Instead, our observations show that the model is also sensitive to the content of syntactically-independent intervening material linearly close to the target verb. 





\section{Discussion} 


Previous NA studies have led \citet{baroni2} to claim that ``the linguistic proficiency of neural networks extends beyond shallow pattern recognition''. Though it is undeniable that BERT does generalize beyond its input and is able to carry out the NA task on the simplest templates, our experiments also suggest that these generalizations can be lexically dependent. 
When naturally occurring lexical patterns are replaced with syntactically well-formed, but meaningless combinations, the model's syntactic ability seems to be heavily compromised, contrary to \citet{goldberg}'s reported results on the \citet{gulordava-etal-2018-colorless} stimuli.

Moreover, most disruption is caused by replacing the words closest to the target within the embedded phrase, that in principle should not affect the agreement relation. These two facts together indicate that some of BERT's syntactic abilities are limited to specific word sequences that the model could have memorized during training, including words that are linearly close but belong to a different embedded phrase or clause.
Furthermore, the fact that the model improves its performance on data it has been trained on (i.e., the \textbf{WIKI} dataset) over other meaningful, unseen  sentences (i.e., the \textbf{M\&L} dataset) is further evidence that at least part of its alleged generalization abilities might be just the effect of memorization.

We can surmise that the model relies on a variety of heuristics acquired during training to approximate syntactic generalizations, in line with \citet{finlayson-etal-2021-causal}, who found two distinct mechanisms to accomplish agreement in Transformer-based architectures. We find that those heuristics can therefore tend to be highly lexicalized, similarly to \citet{newman-etal-2021-refining} who showed that generalization is not systematic by testing a wide range of verbs. This is confirmed by BERT's sensitivity to the main verb when there is no overt complementizer\footnote{cf. sentence type B \textit{no that}}, which prevents it from solving the NA task. This suggests that the model has acquired semi-lexicalized syntactic information about verb subcategorization preferences.

Although BERT's ability to approximate syntactic rules is probably more brittle than previously argued, this should not lead to rejecting its ability to learn natural language grammar. For instance, constructionist approaches \cite{Hoffman:Trousdale:2013} have argued since long against a purely abstract grammar detached from lexical meaning, despite what the data in \ref{ex1} have often been claimed to prove. The alternative view is a grammar consisting of constructions that differ for their level of abstractness and lexicalization. 
BERT's lexically-driven behavior could therefore be consistent with this less abstract conceptions of syntax.
Finally, given previous experiments \cite{laurinavichyute}, we can speculate that humans could also similarly manifest patterns of errors driven by semantic, or lexical interferences from words linearly close to the target. Though such patterns seem to differ between language models and humans \cite{linzen-distinct}, this in turn leads us to questioning our expectations regarding the syntactic abilities of neural language models.

\section{Conclusion}
In this paper, we have shown that BERT's ability to solve the NA task on meaningless sentences strongly depends on the stimuli's syntactic template. While the model is able to perform lexically-independent generalization in simple settings, it fails when the agreement relation crosses an embedded phrase containing an attractor. We further provide insights on this lexical dependence, showing that the model relies mostly on the lexical content at the closest positions to the target of the agreement, though they belong to an independent embedded phrase. 
 
In the future, we want to get a better understanding of the mechanisms underlying the observed syntactic abilities of Transformers, and in particular what makes some heuristics involved to solve a syntactic task lexically dependent. 
A more detailed analysis of the influence played by lexical combinations will help us understand the nature of the heuristics the model uses to solve complex NA cases involving one or more attractors.
Moreover, we wish to compare BERT's predictions with human judgments on our meaningless sentences.


\section{Acknowledgements}
This work was funded in part by the French government under management of Agence Nationale de la Recherche as part of the``Investissements d'avenir" program, reference ANR-19-P3IA-0001 (PRAIRIE 3IA Institute).

\bibliographystyle{acl_natbib}
\bibliography{bibli}

\begin{thebibliography}{15}
\expandafter\ifx\csname natexlab\endcsname\relax\def\natexlab#1{#1}\fi

\bibitem[{Bacon and Regier(2019)}]{bacon}
Geoff Bacon and Terry Regier. 2019.
\newblock \href {http://arxiv.org/abs/1908.09892} {Does {BERT} agree?
  evaluating knowledge of structure dependence through agreement relations}.
\newblock \emph{ArXiv}, abs/1908.09892.

\bibitem[{Baroni(2019)}]{baroni2}
Marco Baroni. 2019.
\newblock \href {http://arxiv.org/abs/1904.00157} {Linguistic generalization
  and compositionality in modern artificial neural networks}.
\newblock \emph{Philosophical Transactions of the Royal Society B}, 375(1791).

\bibitem[{Chomsky(1957)}]{Chomsky:1957}
Noam Chomsky. 1957.
\newblock \emph{Syntactic Structures}.
\newblock Mouton, The Hague.

\bibitem[{Corbett(2003)}]{Corbett2003AgreementTA}
G.~Corbett. 2003.
\newblock Agreement: Terms and boundaries.
\newblock In \emph{The Role of Agreement in Natural Language: TLS 5
  Proceedings}, pages 109--122.

\bibitem[{Devlin et~al.(2019)Devlin, Chang, Lee, and
  Toutanova}]{devlin-etal-2019-bert}
Jacob Devlin, Ming-Wei Chang, Kenton Lee, and Kristina Toutanova. 2019.
\newblock \href {https://doi.org/10.18653/v1/N19-1423} {{BERT}: Pre-training of
  deep bidirectional transformers for language understanding}.
\newblock In \emph{Proceedings of the 2019 Conference of the North {A}merican
  Chapter of the Association for Computational Linguistics: Human Language
  Technologies, Volume 1 (Long and Short Papers)}, pages 4171--4186,
  Minneapolis, Minnesota. Association for Computational Linguistics.

\bibitem[{Finlayson et~al.(2021)Finlayson, Mueller, Gehrmann, Shieber, Linzen,
  and Belinkov}]{finlayson-etal-2021-causal}
Matthew Finlayson, Aaron Mueller, Sebastian Gehrmann, Stuart Shieber, Tal
  Linzen, and Yonatan Belinkov. 2021.
\newblock \href {https://doi.org/10.18653/v1/2021.acl-long.144} {Causal
  analysis of syntactic agreement mechanisms in neural language models}.
\newblock In \emph{Proceedings of the 59th Annual Meeting of the Association
  for Computational Linguistics and the 11th International Joint Conference on
  Natural Language Processing (Volume 1: Long Papers)}, pages 1828--1843,
  Online. Association for Computational Linguistics.

\bibitem[{Goldberg(2019)}]{goldberg}
Yoav Goldberg. 2019.
\newblock \href {http://arxiv.org/abs/1901.05287} {Assessing bert's syntactic
  abilities}.
\newblock \emph{CoRR}, abs/1901.05287.

\bibitem[{Gulordava et~al.(2018)Gulordava, Bojanowski, Grave, Linzen, and
  Baroni}]{gulordava-etal-2018-colorless}
Kristina Gulordava, Piotr Bojanowski, Edouard Grave, Tal Linzen, and Marco
  Baroni. 2018.
\newblock \href {https://doi.org/10.18653/v1/N18-1108} {Colorless green
  recurrent networks dream hierarchically}.
\newblock In \emph{Proceedings of the 2018 Conference of the North {A}merican
  Chapter of the Association for Computational Linguistics: Human Language
  Technologies, Volume 1 (Long Papers)}, pages 1195--1205, New Orleans,
  Louisiana. Association for Computational Linguistics.

\bibitem[{Hoffman and Trousdale(2013)}]{Hoffman:Trousdale:2013}
Thomas Hoffman and Graeme Trousdale, editors. 2013.
\newblock \emph{The Oxford Handbook of Construction Grammar}.
\newblock Oxford University Press, Oxford.

\bibitem[{Lakretz et~al.(2019)Lakretz, Kruszewski, Desbordes, Hupkes, Dehaene,
  and Baroni}]{lakretz-etal-2019-emergence}
Yair Lakretz, German Kruszewski, Theo Desbordes, Dieuwke Hupkes, Stanislas
  Dehaene, and Marco Baroni. 2019.
\newblock \href {https://doi.org/10.18653/v1/N19-1002} {The emergence of number
  and syntax units in {LSTM} language models}.
\newblock In \emph{Proceedings of the 2019 Conference of the North {A}merican
  Chapter of the Association for Computational Linguistics: Human Language
  Technologies, Volume 1 (Long and Short Papers)}, pages 11--20, Minneapolis,
  Minnesota. Association for Computational Linguistics.

\bibitem[{Laurinavichyute and von~der Malsburg(2022)}]{laurinavichyute}
Anna Laurinavichyute and Titus von~der Malsburg. 2022.
\newblock \href {https://doi.org/https://doi.org/10.1111/cogs.13086} {Semantic
  attraction in sentence comprehension}.
\newblock \emph{Cognitive Science}, 46(2):e13086.

\bibitem[{Linzen et~al.(2016)Linzen, Dupoux, and
  Goldberg}]{linzen-etal-2016-assessing}
Tal Linzen, Emmanuel Dupoux, and Yoav Goldberg. 2016.
\newblock \href {https://doi.org/10.1162/tacl_a_00115} {Assessing the ability
  of {LSTM}s to learn syntax-sensitive dependencies}.
\newblock \emph{Transactions of the Association for Computational Linguistics},
  4:521--535.

\bibitem[{Linzen and Leonard(2018)}]{linzen-distinct}
Tal Linzen and Brian Leonard. 2018.
\newblock \href {http://arxiv.org/abs/1807.06882} {Distinct patterns of
  syntactic agreement errors in recurrent networks and humans}.
\newblock \emph{CoRR}, abs/1807.06882.

\bibitem[{Marvin and Linzen(2018)}]{marvin-linzen-2018-targeted}
Rebecca Marvin and Tal Linzen. 2018.
\newblock \href {https://doi.org/10.18653/v1/D18-1151} {Targeted syntactic
  evaluation of language models}.
\newblock In \emph{Proceedings of the 2018 Conference on Empirical Methods in
  Natural Language Processing}, pages 1192--1202, Brussels, Belgium.
  Association for Computational Linguistics.

\bibitem[{Newman et~al.(2021)Newman, Ang, Gong, and
  Hewitt}]{newman-etal-2021-refining}
Benjamin Newman, Kai-Siang Ang, Julia Gong, and John Hewitt. 2021.
\newblock \href {https://doi.org/10.18653/v1/2021.naacl-main.290} {Refining
  targeted syntactic evaluation of language models}.
\newblock In \emph{Proceedings of the 2021 Conference of the North American
  Chapter of the Association for Computational Linguistics: Human Language
  Technologies}, pages 3710--3723, Online. Association for Computational
  Linguistics.

\end{thebibliography}

\newpage

\appendix

\onecolumn

\begin{table*}
\begin{small}
\begin{center}
\begin{tabular}{ cll } 
 \hline
 \textbf{Struct. ID} & \textbf{Structure description} & \textbf{Example} \\
 \hline
 \textbf{A} & Simple agreement & The \textbf{\textcolor{blue}{window}} \textbf{\textcolor{red}{fails}}/*fail \\
 \textbf{B} & In a sentential complement & The prisons insist the \textbf{\textcolor{blue}{surprise}} \textbf{\textcolor{red}{happens}}/*happen \\
 \textbf{C} & Across a prepositional phrase & The \textbf{\textcolor{blue}{gift}} in the \underline{origins} \textbf{\textcolor{red}{reflects}}/*reflect \\
 \textbf{D} & Across a subject relative clause & The \textbf{\textcolor{blue}{passion}} that identifies the \underline{sellers} \textbf{\textcolor{red}{binds}}/*bind \\
 \textbf{E} & In a short verb phrase coordination & The \textbf{\textcolor{blue}{pepper}} falls and \textbf{\textcolor{red}{pulls}}/*pull \\
 \textbf{F} & Across an object relative clause & The \textbf{\textcolor{blue}{bombings}} that the \underline{tune} picks \textbf{\textcolor{red}{flows}}/*flow \\
 \textbf{G} & Within an object relative clause & The rhyme that the \textbf{\textcolor{blue}{elders}} \textbf{\textcolor{red}{need}}/*needs happens \\
 \textbf{H} & Across an object relative clause (\textit{no that}) & The \textbf{\textcolor{blue}{decrees}} the \underline{cage} examine \textbf{\textcolor{red}{happen}}/*happens \\
 \textbf{I} & Within an object relative clause (\textit{no that}) & The lyric the \textbf{\textcolor{blue}{beetles}} \textbf{\textcolor{red}{quote}}/*quotes scores \\
 \hline
\end{tabular}
\caption{\label{generated-examples} Randomly picked examples of generated sentences for each tested structure.}
\end{center}
\end{small}
\end{table*}

\section{Appendix - Data collection}
\label{sec:data-collection}
\subsection{Wikipedia data collection} \label{sec:natural-data} For each of the structures described in \ref{agreement-structures}, we represent the construction by its sequence of lexical categories. We then extract sequences of words from Wikipedia for each of the constructions that match the pattern. To do so, we read Wikipedia linearly and store naturally occurring token sequences that match our constructions, based on the same vocabulary that we use to generate our \textbf{NONCE} sentences, described in \ref{sec:vocabulary}.

\subsection{Data Generation procedure} Generated sentences are built from the sequence of POS-tags describing each construction. We randomly pick one word from our dictionaries at each position of the sequence, as in \cite{gulordava-etal-2018-colorless}. When a noun intervenes between the cue and the target (e.g., in condition C from Table \ref{agreement-structures}), it is systematically assigned a different number from the cue, in order to test attraction effects\footnote{That is whether the model succeeds despite the presence of a distractor noun between the cue and target of the agreement.}. We chose to only use neutral determiners along with possessives to avoid clashes between a noun's and its determiner's numbers. Datasets contain 10000 samples, and for Exp. 2, we reproduced the experiments 10 times for each replacement to produce error bars. Our data is balanced, which means each dataset contains 5000 singulars and 5000 plurals. Randomly picked examples are displayed in \cref{generated-examples}.

\subsection{Data Generation Vocabulary Collection and Preprocessing}
Nouns and verbs were collected from \citet{linzen-etal-2016-assessing}'s dataset. As the NA task setting requires looking at predicted scores for the masked target forms, we only keep verbs for which both forms are present in BERT's vocabulary as an unsplit token. Similarily to Goldberg \cite{goldberg}, we filter out sentences where the target is a present form of the verb 'be' as this verb is too frequent in corpora and is treated differently from other verbs. Our data generation procedure and vocabulary are publicly available at \url{https://github.com/karimlasri/does-bert-really-agree}.

\subsection{Used Vocabulary}
\label{sec:vocabulary}
\paragraph{Determiners and possessives.} `my', `your', `his', `her', `its', `our', `their', `the'
\paragraph{Relativizer/complementizer.} `that'
\paragraph{Nouns.} We use 2636 noun pairs for which both the singular and plural forms are part of BERT's vocabulary.

 \paragraph{Verbs.} We use 444 verb pairs, for which both singular and plural forms are present in BERT's vocabulary.

\paragraph{Stative verbs in Condition B-3.} We use the following stative verbs for the (B-3) condition: 
(`believes', `believe'), (`considers', `consider'), (`doubt', `doubt'), (`hears', `hear'), (`knows', `know'), (`realises', `realise'), (`says', `say'), (`supposes', `suppose'), (`thinks', `think'), (`understands', `understand'), (`wishes', `wish')

\end{document}